\documentclass[letterpaper]{article} 
\usepackage[draft]{aaai25}  
\usepackage{times}  
\usepackage{helvet}  
\usepackage{courier}  
\usepackage[hyphens]{url}  
\usepackage{graphicx} 
\usepackage{natbib}  
\usepackage{caption} 
\usepackage{amssymb}
\usepackage{amsmath}
\usepackage{algorithm}
\usepackage{algorithmic}
\usepackage{multirow}
\usepackage{booktabs}
\usepackage[table]{xcolor}
\usepackage{makecell}
\usepackage{appendix}
\usepackage{newfloat}
\usepackage{listings}
\DeclareCaptionStyle{ruled}{labelfont=normalfont,labelsep=colon,strut=off} 
\floatstyle{ruled}
\newfloat{listing}{tb}{lst}{}
\floatname{listing}{Listing}
\title{Light-T2M: A Lightweight and Fast Model for Text-to-motion Generation}
\author {
    Ling-An Zeng\textsuperscript{\rm 1},
    Guohong Huang\textsuperscript{\rm 1},
    Gaojie Wu\textsuperscript{\rm 1},
    Wei-Shi Zheng\textsuperscript{\rm 1 \rm 2}\thanks{Corresponding Author.}
}
\affiliations {
    \textsuperscript{\rm 1}Sun Yat-sen University\\
    \textsuperscript{\rm 2} Key Laboratory of Machine Intelligence and Advanced Computing, Ministry of Education, China\\
    zenglan3@mail2.sysu.edu.cn, huanggh37@mail2.sysu.edu.cn, wugj7@mail2.sysu.edu.cn, wszheng@ieee.org
    
}

\usepackage{bibentry}

\begin{document}

\newcommand{\todo}[1]{{\color{black}#1}} 
\newcommand{\red}[1]{{\color{black}#1}} 
\newcommand{\lingan}[1]{{\color{black}#1}}

\maketitle

\begin{abstract}
Despite the significant role text-to-motion (T2M) generation plays across various applications, current methods involve a large number of parameters and suffer from slow inference speeds, leading to high usage costs.
To address this, we aim to design a lightweight model to reduce usage costs.
First, unlike existing works that focus solely on global information modeling, we recognize the importance of local information modeling in the T2M task by reconsidering the intrinsic properties of human motion, leading us to propose a lightweight Local Information Modeling Module.
Second, we introduce Mamba to the T2M task, reducing the number of parameters and GPU memory demands, and we have designed a novel Pseudo-bidirectional Scan to replicate the effects of a bidirectional scan without increasing parameter count.
Moreover, we propose a novel Adaptive Textual Information Injector that more effectively integrates textual information into the motion during generation.
By integrating the aforementioned designs, we propose a lightweight and fast model named Light-T2M.
Compared to the state-of-the-art method, MoMask, our Light-T2M model features just \textbf{10\%} of the parameters (4.48M vs 44.85M) and achieves a \textbf{16\%} faster inference time (0.152s vs 0.180s), while surpassing MoMask with an FID of \textbf{0.040} (vs. 0.045) on HumanML3D dataset and \textbf{0.161} (vs. 0.228) on KIT-ML dataset. The code is available at https://github.com/qinghuannn/light-t2m.
\end{abstract}

\begin{figure}[t]
    \centering
    \includegraphics[width=\linewidth]{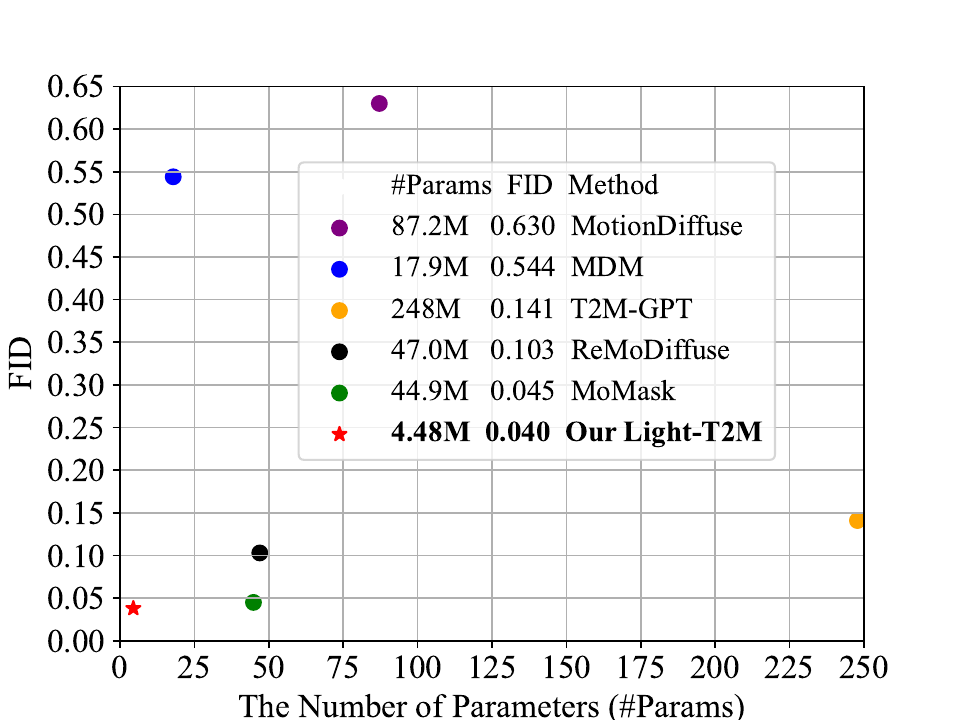}
    \caption{\textbf{Comparison on FID and the number of parameters.} The closer the model is to the origin, the better. Only trainable parameters are calculated. }
    \label{fig:params}
\end{figure}

\section{Introduction}

Text-to-motion (T2M) generation seeks to produce 3D human motion sequences from textual descriptions, playing a crucial role in diverse fields such as film production, video games, and virtual/augmented reality. The broad applicability of T2M has spurred rapid advancements in various methodologies \cite{mld, t2m-gpt, motiondiffuse, momask} in recent years. However, the simplistic and rudimentary design of these models results in high parameter counts and slow inference speeds, which raises the barrier to entry and hampers usability.

Given the imperative to minimize deployment costs and the ubiquitous use of mobile phones, developing lightweight models is essential for the effective implementation of deep learning technologies \cite{liu2024lightweight}. Prior research \cite{efficientnet,deit, mobilestylegan} has focused on creating more efficient models for image classification and generation. In the realm of text-to-motion generation, a streamlined T2M model can substantially lower hardware demands during game and film production and improve the AR/VR and gaming experiences on mobile devices. In this work, we aim to design a lightweight T2M model.

To design a lightweight T2M model, we have recognized the overlooked importance of local information modeling in T2M, ignored by recent methods, and use this to guide model design. Intuitively, the essence of generating realistic human motion lies in the smooth transition between adjacent movements, indicating the importance of local information modeling. However, recent existing methods focus only on the global information interaction modeled by Transformer \cite{vaswani2017attention}. Thus, we design a lightweight Local Information Modeling Module (LIMM) using the lightweight convolution network. By appropriately replacing some global information modeling layers with our LIMM, we significantly reduce the number of parameters without sacrificing performance.

Secondly, we introduce Mamba \cite{mamba,mamba2} into the T2M task to reduce \lingan{the number of parameters} and GPU memory requirements, and design a novel Pseudo-bidirectional Scan to achieve bidirectional scanning without adding extra parameters. \lingan{By replacing the Transformer heavily relied upon in previous methods with Mamba, we further streamline resource usage.} Besides, since the single-direction scan in Mamba, intended for causal sequences, is inadequate for global information modeling, a simple yet effective Pseudo-bidirectional Scan is designed to achieve a bidirectional scan. In detail, we feed the concatenation of the reverse sequence and original sequence into Mamba and, during a single-direction scan, each element in the original sequence can see the information of the element on its left or right. In this way, we achieve the effect of a bidirectional scan without increasing parameter count.

Furthermore, different from existing T2M works that rely on token mixing in self/cross-attention mechanisms to integrate textual information into motion frames, we propose a novel Adaptive Textual Information Injector (ATII) that employs a more effective method for embedding textual information during motion generation. Inspired by the gating mechanism in LSTM \cite{lstm}, our ATII adaptively extracts segment-aware semantics from the text embedding token derived from CLIP \cite{clip}, and incorporates it into corresponding motion segments. Specifically, for each motion segment, the ATII predicts channel-wise weights for the text embedding token and uses these weights to modulate the text token. Subsequently, the modulated text token and the motion segment token are combined in a fusion layer, enabling adaptive injection of textual information. 

By integrating the aforementioned designs, we have developed Light-T2M, a lightweight and fast model tailored for text-to-motion generation. As shown in Figure \ref{fig:params}, Light-T2M employs only 4.48M parameters—approximately \textbf{10\%} of those used by the state-of-the-art method, MoMask \cite{momask}. Despite its smaller size, it excels, achieving an FID of 0.040 (vs. 0.045 of MoMask) on the HumanML3D dataset and 0.161 (vs. 0.228) on the KIT-ML dataset. Owing to its efficient design, Light-T2M also delivers a \textbf{16\%} faster inference time (0.152s vs. 0.180s). Thanks to its lightweight design and fast inference speed, our Light-T2M reduces the cost of use and paves the road for the widespread use of text-to-motion generation. In summary, our main contributions can be summarized as follows:
\begin{itemize}
    \item We introduce  Light-T2M, a lightweight model for text-to-motion generation, aiming to reduce the cost of use and pave the road for the widespread use of text-to-motion generation.
    \item We propose a novel Adaptive Textual Information Injector to inject the textual information into the motion during generation, which is an efficient and effective way to control the generated motion through text.
    \item We realize the importance of local information modeling and introduce Mamba into the text-to-motion task, both reducing the number of model parameters and speeding up inference time, along with a novel Pseudo-bidirectional Scan to achieve the effect of the bidirectional scan without increasing parameters.
\end{itemize}

\begin{figure*}[ht]
    \centering
    \includegraphics[width=\linewidth]{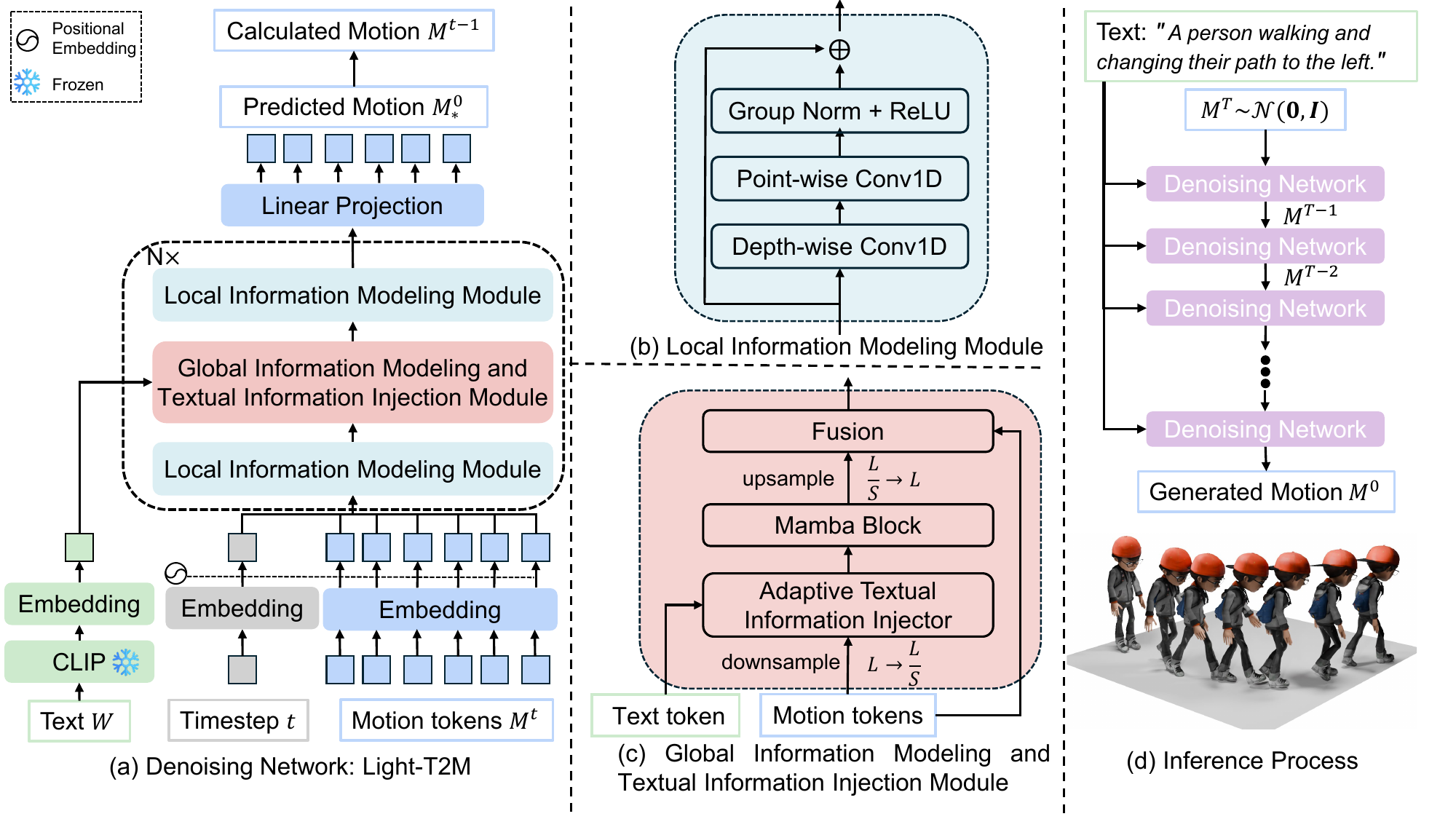}
    \vspace{-0.3cm}
    \caption{\textbf{Overview of our Light-T2M.} (a) Our Light-T2M consisting of $N$ basic blocks aims to predict $M_*^0$, and then $M^{t-1}$ can be calculated via Eq. \ref{eq:2}. (b) The structure of our lightweight Local Information Modeling Module. (c) The motion is downsampled to obtain segments containing local semantic information. Next, a novel Adaptive Textual Information Injector and a Mamba Block are adopted to adaptively inject semantics into each segment and model the global information, respectively. The upsampled motion and the original motion are fused by a fusion layer. (d) The overview of the inference process.}
    \label{fig:overview}
\end{figure*}

\section{Related Work}

\noindent\textbf{Text-to-motion Generation.} 
Early works such as \cite{GhoshCOTS21, plappert2018learning, yamada2018paired} utilize encoder-decoder RNN models to achieve text-to-motion generation. Subsequent studies \cite{t2m, temos, attt2m} adopt a variational autoencoder (VAE) \cite{vae} to encode motion into a latent distribution and transform text into the target latent distribution. Following the success of diffusion models, most recent works \cite{flame, motiondiffuse, mdm, m2dm, priormdm, omnicontrol,HanPDRSX24} resort to diffusion models and build the network based on Transformer to achieve text-driven motion generation. Some researchers \cite{mld, bridege} design diffusion models to predict motion in latent spaces, while others \cite{t2m-gpt, motiongpt} develop autoregressive models to sequentially generate motion frames. A few studies \cite{fg-t2m, remodiffuse} propose specially designed modules instead of directly using Transformers. Recently, Guo et al. \cite{momask} treated the target motion as masked tokens and developed a Masked Transformer to progressively predict all masked tokens. However, due to simplistic and rudimentary architectural designs, these methods require a large number of parameters. In contrast, we aim to design a lightweight model based on our well-designed modules and proposed Adaptive Textual Information Injector.

\vspace{0.05cm}
\noindent\textbf{Lightweight Network.}
Reducing computational costs and enhancing storage capabilities have elevated the importance of lightweight network design in deep learning \cite{lightsurvey}. Projects like the MobileNet series \cite{mobilenets, mehta2022mobilevit, mobileformer, MobileNetV2} demonstrate manual construction of lightweight models informed by prior knowledge. MobileNet, for instance, features a streamlined architecture with depth-wise separable convolutions for creating efficient neural networks \cite{mobilenets}. Techniques like neural architecture search aim to automate model development while minimizing human input \cite{pslt, darts}, and neural network compression offers additional routes to model efficiency \cite{autoprune, quantsurvey, yan2024dialoguenerf}. Yet, the trade-offs between performance and parameters in these methods often lead to suboptimal results. In this work, we tailor our model to meet the unique demands of the T2M task without sacrificing performance.

\vspace{0.05cm}
\noindent\textbf{State Space Models.}
The Structured State-Space Sequence (S4) model is attractive for its linear scaling with sequence length, prompting further exploration. Recently, Mamba \cite{mamba, mamba2}, leveraging a data-dependent SSM layer, has outperformed Transformers in large-scale data, demonstrating its potential for sequence modeling. Following Mamba's success, numerous studies have investigated its application in specific fields. For instance, Vision Mamba \cite{vimamba} and VMamba \cite{vmamba} have developed unique scanning techniques for 2D image processing. In our work, we aim to extend Mamba's advantages to the T2M task by designing a model that is both lightweight and fast.

\section{Methodology}

\subsection{Problem Formulation}
Given a text $W=\{w_i\}$ describing motion, our goal is to generate a corresponding 3D human pose sequence $M=\{m_i\}$ of length $L$. According to existing works \cite{t2m, momask}, each 3D human pose is represented by a $D_m$-dimensional vector, which includes the root angular velocity along the Y-axis, root linear velocities on the XZ-plane, root height, and local joint positions, velocities, and rotations relative to the root space. For more detailed information on human pose representation, please see T2M \cite{t2m}.

\subsection{Overview of Light-T2M}
Figure \ref{fig:overview} illustrates the architecture of our Light-T2M, which consists of $N$ blocks. Each block contains two Local Information Modeling Modules and one Global Information Modeling and Textual Information Injection Module. The former modules model local information within each motion frame's local window, ensuring the consistency of adjacent frames—a critical aspect of generating realistic human motion. The latter module manages global information across all frames and injects textual information, ensuring coherence throughout the sequence and alignment with the given text. Additionally, a fixed CLIP \cite{clip} model and three embedding layers extract text features and harmonize various features within the same space.

\noindent \textbf{Training.} As shown in Figure \ref{fig:overview}, our Light-T2M is a diffusion-based model. 
Following \cite{ddpm}, we define a Markov chain of $T$ diffusion steps to slowly add random noise $\epsilon \sim \mathcal{N}(\mathbf{0},\mathbf{I})$ to motion. Given a motion $M$, and the diffusion time-step $t\in[1, T]$, our denoising network (Light-T2M) $\phi_\theta(\cdot)$ is trained via the following training objective:
\begin{gather}
    \mathcal{L} = \mathbb{E}_{M^0,t,\epsilon}|| M^0 - \phi_\theta(M^t, t, c) ||^2 , \\
    M^t = \sqrt{\bar{\alpha}_t}M^0 + \sqrt{1-\bar{\alpha_t}}\epsilon, \label{eq:2}
\end{gather}
where $M^0=M$, $c$ is the condition text, and a predefined $\alpha_t$ aims to control the variance. While predicting $M^0$ and predicting noise $\epsilon$ are mathematically equivalent, we find that predicting $M^0$ yields better performance in the T2M task. To both learn conditioned or unconditioned generation during training, we randomly set $c=\varnothing$ with a probability $\tau$. 

\noindent \textbf{Inference.} During generation, we reverse the diffusion process to construct desired data samples from the noise $M^T \sim \mathcal{N}(\mathbf{0},\mathbf{I})$. For each time step $t$ and noise motion $M^t$, we first use the trained Light-T2M to predict the conditional sample $\hat{M}^0_*$ and unconditional sample $\bar{M}^0_*$ by setting $c$ as the given text $W$ and $\varnothing$. Using $M^t$, $\hat{M}^0_*$, and $\bar{M}^0_*$, we can calculate the conditional noise $\epsilon_c$ and unconditional noise $\epsilon_u$ via Equation \ref{eq:2}. Finally, we adopt \textit{classifier-free guidance} method \cite{cfg} for conditional generation:
\begin{equation}
    \hat{\epsilon} = (1 + s) \cdot \epsilon_c - s \cdot \epsilon_u,
\end{equation}
where the guidance scale $s$ controls the strength of the condition. The obtained noise $\hat{\epsilon}$ is then used to calculate $M^{t-1}$. More details are shown in the Appendix.

\subsection{Learning to Model Local Information}
Recognizing that modeling the interaction of location information across adjacent frames can enhance the smoothness of generated motion and be implemented efficiently, we have integrated location information modeling into the T2M task. We believe that strategically replacing some global information modeling layers with location information modeling layers can significantly reduce the parameter count without sacrificing performance.

Inspired by the success of CNN in local information modeling, we design our Local Information Modeling Modules (LIMM) based on lightweight 1D convolution layers. As shown in Figure \ref{fig:overview}.b, to reduce the number of parameters and FLOPs, our LIMM mainly consists of a 1D point-wise convolution $f^p(\cdot)$ and a 1D depth-wise convolution $f^d(\cdot)$. Formally, our LIMM can defined as:
\begin{equation}
    \hat{X} = X + ReLU(Norm(f^p(f^d(X)))),
\end{equation}
where $X\in\mathbb{R}^{D\times L}$ and $Norm(\cdot)$ are the input motion and the Group Normalization \cite{groupcnn}. 

\begin{figure}
    \centering
    \includegraphics[width=0.8\linewidth]{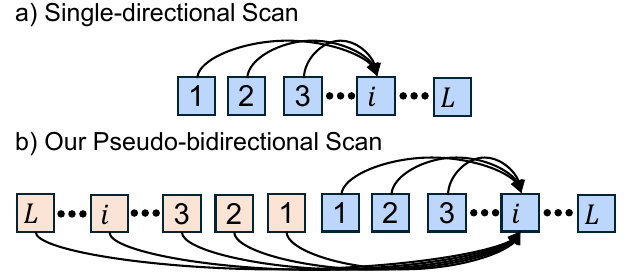}
    \caption{In our pseudo-bidirectional scan, each element in the original sequence can obtain the information from elements originally on its right, achieving the effect of bidirectional scanning without increasing parameters.}
    \label{fig:scan}
\end{figure}

\subsection{Learning to Model Global Information}
Beyond modeling local information, modeling global information and injecting textual information are crucial for text-driven motion generation. The former ensures rationality and semantic consistency across the overall motion, while the latter aligns the motion semantically with the provided text. Given their interrelated nature, we have consolidated global information modeling and textual information injection into a single module, namely the Global Information Modeling and Textual Information Injection Module.

As shown in Figure \ref{fig:overview}.c, we first downsample the input motion sequence in the time dimension by a ratio of $S$ \lingan{via a lightweight point-wise convolution layer}. We then inject textual information into each segment using our proposed Adaptive Textual Information Injector (ATII). Next, we utilize Mamba to model the global information using our proposed pseudo-bidirectional scan. Subsequently, we upsample the segments back to their original length through repeated padding. Finally, we fuse the original motion sequence $X$ with the upsampled motion sequence $\bar{X}$ using a fully connected layer $h(\cdot)$:
\begin{equation}
    \hat{X} = h(X + \bar{X}).
\end{equation}

\noindent\textbf{Benefits of Downsample.} 
We note that the downsample operation has two benefits. Firstly, the semantic information in a single frame is typically weak, making it challenging to extract corresponding semantics from the given text. Furthermore, modeling global information at the frame level is inefficient. Moreover, downsampling reduces the length of the motion sequence from $L$ to $\frac{L}{S}$, significantly decreasing the computational load of subsequent operations.

\noindent\textbf{Pseudo-bidirectional Scan.} 
Since the original single-direction scan used in Mamba limits information exchange, we propose a simple yet effective method to achieve a bidirectional scan without adding extra parameters. As shown in Figure \ref{fig:scan}, we reverse the input motion sequence and feed both the reversed $X^r$ and the original $X^o$ sequences into Mamba. Then, during the sequential scan, each element in the original sequence $X^o$ can access information from adjacent elements on either side. After scanning, only the original sequence $X^o$ is retained. With this approach, we achieve the effect of a bidirectional scan, as in \cite{vimamba}, and a cross-scan, as in \cite{vmamba}, without increasing the number of parameters. We conduct experiments that demonstrate the superior efficacy of our pseudo-bidirectional scan.

\begin{figure}
    \centering
    \includegraphics[width=0.6\linewidth]{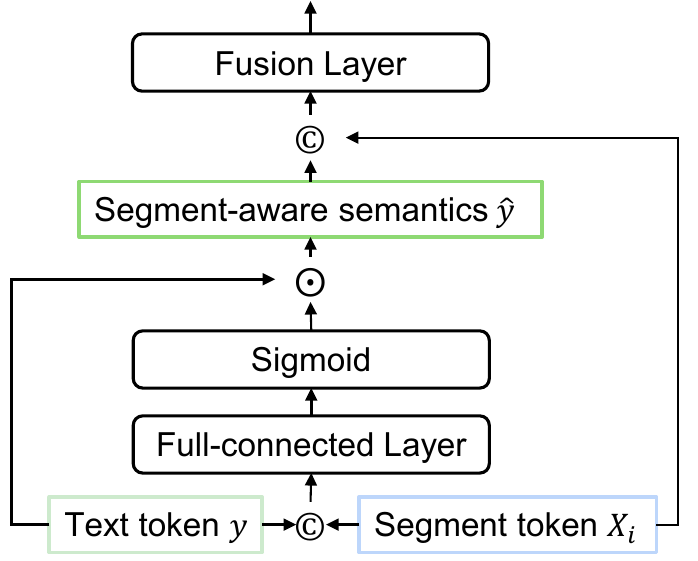}
    \caption{\textbf{Illustration of our Adaptive Textual Information Injector.} $\odot$ and $\copyright$ denote dot product and concatenation, respectively. }
    \label{fig:atii}
\end{figure}

\begin{table*}[ht]
\centering
\setlength\tabcolsep{1mm}
\begin{tabular}{ccccccccc}
\bottomrule
 \multirow{2}{*}{Methods} & \multirow{2}{*}{\#Params} & \multirow{2}{*}{AIT$\downarrow$} & \multirow{2}{*}{FID$\downarrow$}  & \multicolumn{3}{c}{R-Precision$\uparrow$} & \multirow{2}{*}{MM. Dist.$\downarrow$} & \multirow{2}{*}{MM. $\uparrow$} \\ \cline{5-7}
   & & & & Top1 & Top2 & Top3 & & \\ \toprule
\rowcolor{gray!30} \multicolumn{9}{l}{\textit{On the HumanML3D dataset.}}\\ \hline 
TM2T (\citeauthor{tm2t}) & 41.05M & 0.680s & $1.501^{\pm.017}$ & $0.424^{\pm.003}$ & $0.618^{\pm.003}$ & $0.729^{\pm.002}$ & $3.467^{\pm.011}$ & $\underline{2.424}^{\pm.093}$ \\
T2M (\citeauthor{t2m}) & 31.16M & \textbf{0.038s} & $1.087^{\pm.021}$ & $0.455^{\pm.003}$ & $0.636^{\pm.003}$ & $0.736^{\pm.002}$ & $3.347^{\pm.008}$ & $2.219^{\pm.074}$\\
MDM (\citeauthor{mdm}) & 17.88M & 14.32s  & $0.544^{\pm.044}$ & - & - & $0.611^{\pm.007}$ & $5.566^{\pm.027}$ & $\textbf{2.799}^{\pm.072}$ \\
MLD (\citeauthor{mld}) & 26.38M & 0.250s & $0.473^{\pm.013}$ & $0.481^{\pm.003}$ & $0.673^{\pm.003}$ & $0.772^{\pm.002}$ & $3.196^{\pm.010}$  & $2.413^{\pm.079}$                \\
MotionDiffuse (\citeauthor{motiondiffuse}) & 87.15M & 17.36s & $0.630^{\pm.001}$ & $0.491^{\pm.001}$ & $0.681^{\pm.001}$                & $0.782^{\pm.001}$                & $3.113^{\pm.001}$                      & $1.553^{\pm.042}$                \\
T2M-GPT  (\citeauthor{t2m-gpt}) & 247.80M & 0.434s & $0.141^{\pm.005}$ & $0.492^{\pm.003}$ & $0.679^{\pm.002}$ & $0.775^{\pm.002}$ & $3.121^{\pm.009}$                      & $1.831^{\pm.048}$                \\
ReMoDiffuse (\citeauthor{remodiffuse}) & 46.97M & 0.475s & $0.103^{\pm.004}$ & $0.510^{\pm.005}$ & $0.698^{\pm.006}$ & $\underline{0.795}^{\pm.004}$                & $\underline{2.974}^{\pm.016}$                      & $1.795^{\pm.043}$                \\
MoMask (\citeauthor{momask}) & 44.85M & 0.180s & $0.045^{\pm.002}$ & $\textbf{0.521}^{\pm.002}$ & $\textbf{0.713}^{\pm.002}$ & $\textbf{0.807}^{\pm.002}$ & $\textbf{2.958}^{\pm.008}$ & $1.241^{\pm.040}$                \\ \hline
\textbf{Our Light-T2M} & \textbf{4.48M} & \underline{0.151s} & $\textbf{0.040}^{\pm{.002}}$ & $\underline{0.511}^{\pm.003}$ & $\underline{0.699}^{\pm.002}$ & $\underline{0.795}^{\pm.002}$ & $3.002^{\pm.008}$ & $1.670^{\pm.061}$ \\ \toprule

\rowcolor{gray!30} \multicolumn{9}{l}{\textit{On the KIT-ML dataset.}}\\ \hline 
TM2T (\citeauthor{tm2t}) & - & - & $3.599^{\pm.153}$ & $0.280^{\pm.005}$ & $0.463^{\pm.006}$ & $0.587^{\pm.005}$ & $4.591^{\pm.026}$ & $\textbf{3.292}^{\pm.081}$ \\
T2M (\citeauthor{t2m}) & - & - & $3.022^{\pm.107}$ & $0.361^{\pm.005}$ & $0.559^{\pm.007}$ & $0.681^{\pm.007}$ & $3.488^{\pm.028}$ & $2.052^{\pm.107}$\\
MDM (\citeauthor{mdm}) & - & - & $0.497^{\pm.021}$ & - & - & $0.396^{\pm.004}$ & $9.191^{\pm.022}$ & $1.907^{\pm.214}$ \\
MLD (\citeauthor{mld}) & - & - & $0.404^{\pm.027}$ & $0.390^{\pm.008}$ & $0.609^{\pm.008}$ & $0.734^{\pm.007}$ & $3.204^{\pm.027}$  & $\underline{2.192}^{\pm.071}$                \\
MotionDiffuse (\citeauthor{motiondiffuse}) & - & - & $1.954^{\pm.062}$ & $0.417^{\pm.004}$ & $0.621^{\pm.004}$                & $0.739^{\pm.004}$                & $2.958^{\pm.005}$                      & $0.730^{\pm.013}$                \\
T2M-GPT  (\citeauthor{t2m-gpt}) & - & - & $0.514^{\pm.029}$ & $0.416^{\pm.006}$ & $0.627^{\pm.006}$ & $0.745^{\pm.006}$ & $3.007^{\pm.023}$                      & $1.570^{\pm.039}$                \\
ReMoDiffuse (\citeauthor{remodiffuse})  & - & - & $\textbf{0.155}^{\pm.006}$ & $0.427^{\pm.014}$ & $0.641^{\pm.004}$ & $0.765^{\pm.055}$                & $2.814^{\pm.012}$                      & $1.239^{\pm.028}$                \\
MoMask (\citeauthor{momask}) & - & - & $0.204^{\pm.011}$ & $\underline{0.433}^{\pm.007}$ & $\underline{0.656}^{\pm.005}$ & $\underline{0.781}^{\pm.005}$ & $\underline{2.779}^{\pm.022}$ & $1.131^{\pm.043}$                \\ \hline
\textbf{Our Light-T2M} & - & - & $\underline{0.161}^{\pm.009}$ & $\textbf{0.444}^{\pm.006}$ & $\textbf{0.670}^{\pm.007}$ & $\textbf{0.794}^{\pm.005}$ & $\textbf{2.746}^{\pm.016}$ & $1.005^{\pm.036}$\\ \toprule
\end{tabular}
\vspace{-0.1cm}
\caption{\textbf{Quantitative evaluation on the HumanML3D and KIT-ML test set.} Following previous works, we replicated the experiment 20 times to calculate the average results, presented with a 95\% confidence interval (denoted by$\pm$). The best result is bolded and the second is underlined. \lingan{Average Inference time (AIT) is calculated from the average across 100 samples using the same RTX 3090Ti GPU.} }
\label{tab:hml}
\end{table*}

\subsection{Injecting Textual Information into Motion}
Unlike most existing T2M works that rely on self/cross-attention in Transformers for conditional generation, we propose a novel Adaptive Textual Information Injector (ATII) to inject textual information into the motion during generation more effectively. 

As shown in Figure \ref{fig:atii}, the concatenation of the text embedding token $y$ and a segment token $X_i$ is fed into a fully connected layer $g^c(\cdot)$ followed by a Sigmoid function to obtain channel-wise weights for the text embedding token, inspired by the gating mechanism in LSTM \cite{lstm}, and then the channel-wise weights are used to reweight the text token:
\begin{equation}
    \hat{y} = Sigmoid(g^c(X_i, y)) \odot y, 
\end{equation}
where $\odot$ denotes the point-wise production. Thus, the obtained feature $\hat{y}$ contains the segment-aware semantics. After that, another fully connected layer $g^f(\cdot)$ is adopted to fuse obtained feature $\hat{y}$ and the segment token $X_i$:
\begin{equation}
    \hat{X_i} = g^f(X_i, \hat{y}).
\end{equation}
In this way, we successfully extract adaptive semantics from the text token for each motion segment and then inject such information into the motion segment. 

\begin{figure*}[ht]
    \centering
    \setlength{\tabcolsep}{0cm}
    \begin{tabular}{cccc}
    \toprule
    \makebox[0.23\textwidth][c]{\textbf{Light-T2M (Ours)}} & \makebox[0.24\textwidth][c]{MoMask} & \makebox[0.25\textwidth][c]{T2M-GPT} & \makebox[0.23\textwidth][c]{MotionDiffuse} \\ \hline
    \multicolumn{4}{c}{A person \textbf{walks forward} then \textbf{sits down}.} \\
    \multicolumn{4}{c}{\includegraphics[width=\linewidth]{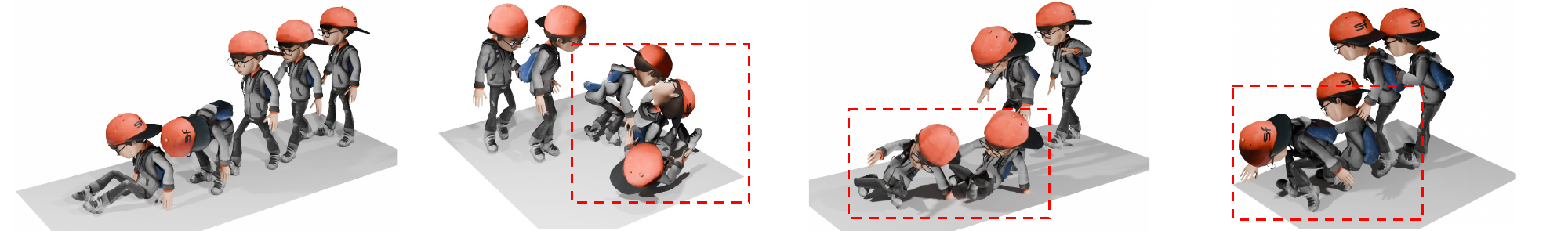}} \vspace{-0.05cm }\\ \hline
    \multicolumn{4}{c}{A person \textbf{kneels down onto the floor}.} \\
    \multicolumn{4}{c}{\includegraphics[width=\linewidth]{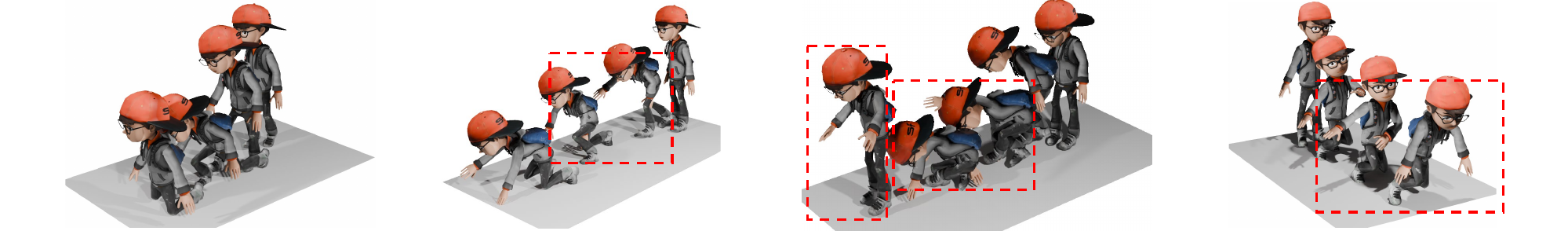}} \vspace{-0.05cm }\\ \hline
    \multicolumn{4}{c}{A person \textbf{jumps up} then waits for a bit and then \textbf{walks forwards}.} \\
    \multicolumn{4}{c}{\includegraphics[width=\linewidth]{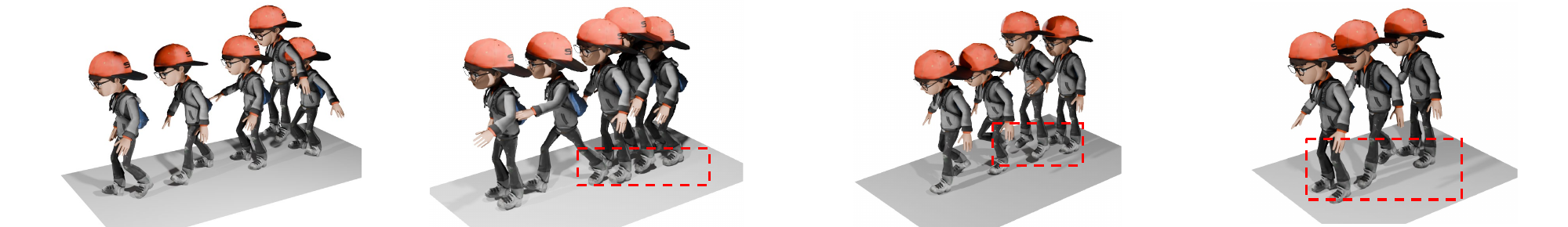}} \vspace{-0.05cm }\\ \hline
    \multicolumn{4}{c}{A person \textbf{walks} in a line starting first to the \textbf{right}, then \textbf{forward}, then \textbf{left} for a distance, and then \textbf{forward} again.} \\
    \multicolumn{4}{c}{\includegraphics[width=\linewidth]{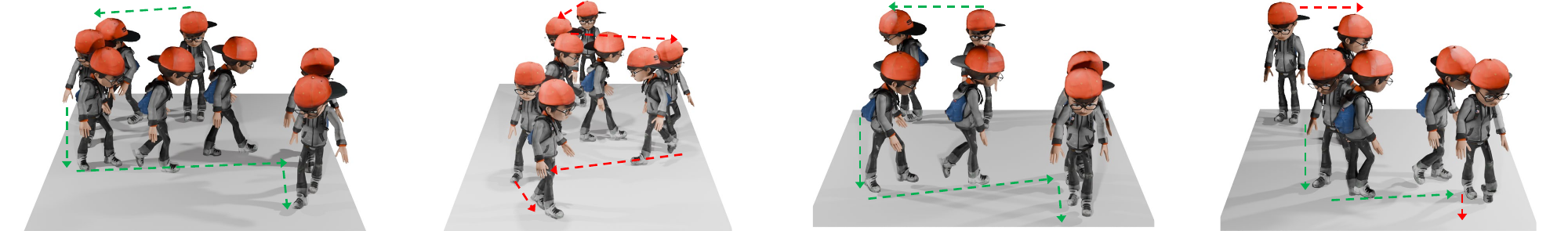}} \vspace{-0.05cm }\\ 
    \toprule
    \end{tabular}
    \caption{\textbf{Qualitative comparisons on the HumanML3D dataset.} The areas highlighted in red indicate where the generated content does not correspond to the given text or where there are issues such as limb distortion. We also use dashed lines to display the character's movement path, with green and red indicating whether it corresponds to the given text, respectively.}
    \label{fig:vis}
\end{figure*}
    

\section{Experiments}\label{sec:exp}

\subsection{Datasets and Evaluation Metrics}
\noindent\textbf{Datasets.} We conduct experiments on two most common public text-motion datasets, i.e., the HumanML3D dataset \cite{t2m} and the KIT-ML dataset \cite{kit-ml}. The \textbf{HumanML3D} dataset, constructed based on the HumanAct12 \cite{humanact12} and AMASS \cite{amass}, contains 14,616 motion sequences and 44,970 text descriptions. This dataset contains diverse activities including exercising, sports, and acrobatics. The \textbf{KIT-ML} dataset is smaller and contains 3,911 motion sequences and 6,278 text descriptions. For both datasets, the preprocessing procedure and the train-test-validation split remain consistent with \cite{t2m}.

\noindent\textbf{Evaluation Metrics.}
We strictly adhere to the evaluation pipeline and adopt the same evaluation metrics used in \cite{t2m}. Each motion-text pair is fed into a pretrained network to extract feature embeddings and then measured by the following metrics: (1) \textit{Frechet Inception Distance (FID)} measures the overall motion quality by calculating the similarity between features of generated and ground truth motions. (2) \textit{R-Precision} and \textit{multimodal distance (MM. Dist.)} assess the semantic alignment between the input text and the generated motions. (3) \textit{Multimodality (MM)} aims to assess the diversity of motions generated from the same text. Following recommendations from MoMask \cite{momask}, FID and R-Precision serve as the primary performance metrics, and the diversity metric is not included.

\subsection{Implementation Details}
The max diffusion step $T$ is 1000 and the linearly varying variances $\beta_t$ range from $10^{-4}$ to $10^{-2}$. During inference, we adopt UniPC \cite{unipc} with 10 time steps for the fast sampling. The number of blocks $N$, the hidden dim $D$, and the downsampling factor $S$ are 4, 256, and 8, respectively. The guidance scale $s$ and the text dropout ratio $\tau$ are set to 4 and 0.2, respectively. Our Light-T2M is optimized by Adamw \cite{adamw} with a learning rate of 2e-4, a cosine annealing schedule, and a batch size of 256 on 2 RTX 3090Ti GPUs. Light-T2M is trained with 3000/5000 epochs on the HumanML3D/KIT-ML datasets. More details are shown in the Appendix.

\subsection{Comparison with State-of-the-arts}
Since the model parameters are nearly identical across both datasets and some works do not provide pretrained models on the KIT-ML dataset, we report only the number of trainable parameters and the inference time for the HumanML3D dataset. Note that for models utilizing VAE or VQVAE, the parameters of these architectures are included in the count. Due to limited space, we present the performance of state-of-the-art methods and a part of the works in the table.

\vspace{0.1cm}
\noindent \textbf{Quantitative Comparison.}
The quantitative results are shown in Table \ref{tab:hml}. Compared to MoMask, our Light-T2M only uses 10\% of trainable parameters (4.48M vs 44.85M) and achieves a 16\% faster inference time (0.152s vs 0.180s), while outperforming MoMask in FID on both datasets. In terms of R-Precision and multimodal distance, Light-T2M performs better on the KIT-ML dataset but falls short on the HumanML3D dataset. Regarding the suboptimal FID on the KIT-ML dataset, ReMoDiffuse utilizes complex data retrieval from a large database to enhance motion quality, as noted in MoMask. These quantitative results confirm that Light-T2M not only reduces usage costs but also improves the feasibility of deployment on mobile devices.

\vspace{0.1cm}
\noindent \textbf{Qualitative Comparison.}
Figure \ref{fig:vis} shows a qualitative comparison with previous works. MotionDiffuse fails to generate poses for actions such as ``sit down'' and ``knees down'', and it does not accurately interpret the concept of direction. T2M-GPT, while generating semantically correct motions, often produces poses with noticeable limb distortions. The motions generated by MoMask do not align well with the given text and also exhibit limb distortions. In contrast, our model consistently generates correct and reasonable motion poses that accurately reflect the given text and understand directional concepts. Overall, the motion quality produced by our model exceeds that of the existing works.

\begin{figure}[t]
    \centering
    \includegraphics[width=0.8\linewidth]{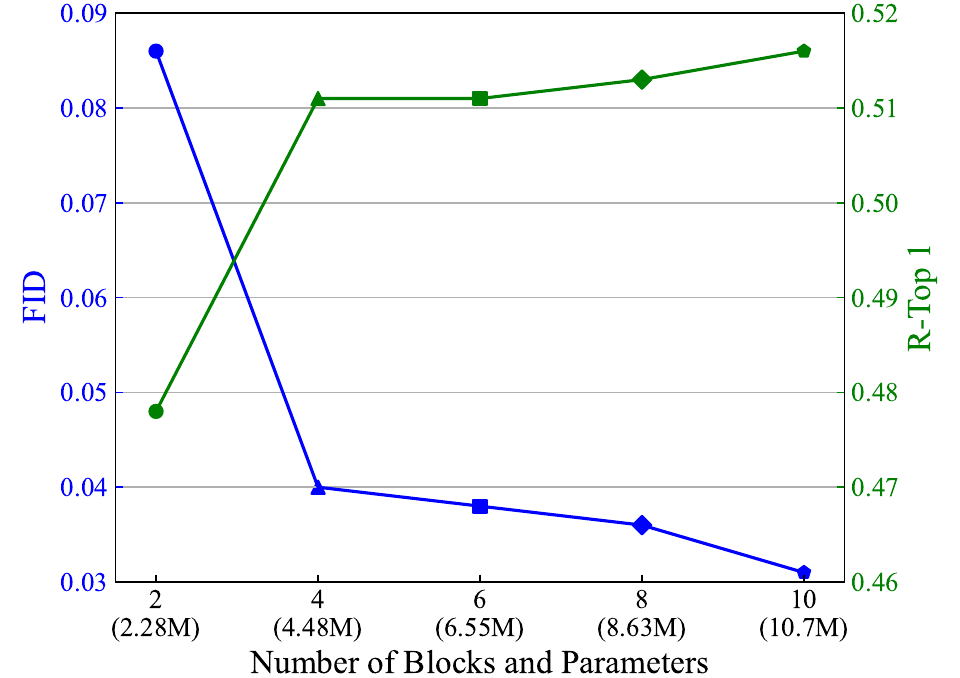}
    \caption{\textbf{Impact of The Number of Parameters.}}
    \label{fig:layer}
\end{figure}

\subsection{Ablation Studies}

All ablation studies are conducted on the HumanML3D dataset and the confidence intervals are omitted for simplicity. More experiments are shown in the Appendix.

\vspace{0.1cm}
\noindent\textbf{Impact of the Number of Parameters.}
As shown in Figure \ref{fig:layer}, we evaluate the model's performance using different numbers of basic blocks shown in Figure \ref{fig:overview}. As the number of modules increases, so does the model's performance. However, while the parameter count rises by 139\% when the number of modules increases from 4 to 10, the FID improvement is only 22.3\%. Consequently, the model with 4 modules is selected as the final configuration. Surprisingly, our Light-T2M with just 2 basic blocks, comprising 2.28M parameters, achieves an FID of 0.086 and an R-Top 1 of 0.478, demonstrating the method's effectiveness and efficiency.

\vspace{0.1cm}
\noindent\textbf{Analysis of Model Design.}
Table \ref{tab:model} shows the performance when replacing basic blocks in our Light-T2M with other basic blocks.
Notably, when using alternate basic blocks, the concatenation of frame tokens with the text token is processed through the global information layer for textual information injection. From this data, we draw several conclusions: (1) The comparisons between ``T'', and ``LTL'' highlight the significance of local information modeling. (2) The comparisons among ``LTL'', ``LML'', and ``LM$^*$L'' underscore the efficiency of the original Mamba and the value of the bidirectional scan. (3) The contrasts between `LTL'', ``LML'', and ``LGL'' demonstrate that our Light-T2M achieves considerable performance enhancements with only a modest increase in parameters. These findings validate our claims made in the introduction. Please refer to the Appendix for more details about the model design.

\vspace{0.1cm}
\noindent\textbf{Comparison of Different Scan Methods.}
We evaluate the performance when replacing our pseudo-bidirectional scan with different scan methods. As shown in the upper part of Table \ref{tab:scan}, the significant performance gap between the single-directional scan and the bidirectional scan underscores the necessity of bidirectional information modeling. Furthermore, compared to the bidirectional scan in \cite{vimamba}, our pseudo-bidirectional scan achieves superior performance without increasing the number of parameters.

\vspace{0.1cm}
\noindent\textbf{Analysis of Adaptive Textual Information Injector (ATII).} 
The lower part of Table \ref{tab:scan} shows the experimental analysis of our ATII.  Note that when removing our ATII, the concatenation of frame tokens with the text token is fed into the global information layer for textual information injection.  \lingan{Besides, in experiments involving the removal of our ATII or the gating mechanism, the downsampling and upsampling processes are retained.} After removing our ATII, the FID and R-Top 1 dropped significantly. When removing the gating mechanism, the performance will also decrease. These results validate the effectiveness of our ATII.

\begin{table}[]
\centering
\setlength\tabcolsep{1.5mm}
\begin{tabular}{ccccc}
\bottomrule
Block Design & \#Params & FID$\downarrow$  & R-Top1$\uparrow$ & R-Top3$\uparrow$ \\ \toprule
T & \underline{3.49M} & 0.214 & 0.437 & 0.727\\
LTL & 4.03M & 0.160 & 0.469 & 0.763 \\
LML & $\textbf{2.63M}$ & 0.165 & 0.456 & 0.736\\
LM$^*$L & 4.38M & $\underline{0.114}$ & $\underline{0.501}$ & $\underline{0.783}$\\
\rowcolor{gray!30} LGL (Ours) & 4.48M & $\textbf{0.040}$ & $\textbf{0.511}$ & $\textbf{0.795}$ \\ \toprule
\end{tabular}
\caption{\textbf{Analysis of Model Design.} We evaluate the performance when replacing basic blocks in Light-T2M with other basic blocks. ``T'' denotes one Transformer encoder layer. ``L'' denotes our Local Information Modeling Module. ``G'' denotes our Global Information Modeling and Injection Textual Information Module. ``M'' and ``M$^*$'' denote an original Mamba block and a BiMamba block in \cite{vimamba}.}
\label{tab:model}
\end{table}

\begin{table}[]
\centering
\setlength\tabcolsep{1.5mm}

\begin{tabular}{ccccc}
\bottomrule
 & \#Params & FID$\downarrow$  & R-Top1$\uparrow$ & R-Top3$\uparrow$\\ \toprule
SDS & \textbf{4.48M} & 0.058 & 0.495 & 0.781\\
BDS & 4.66M & 0.042 & 0.509 & 0.793\\
\rowcolor{gray!30} PBDS (Ours) & \textbf{4.48M} & $\textbf{0.040}$ & $\textbf{0.511}$ & $\textbf{0.795}$  \\ \hline \hline
w/o ATII & \textbf{3.43M} & 0.102 & 0.489 & 0.778\\
w/o gating  & 3.95M & 0.074 & 0.497 & 0.781 \\
\rowcolor{gray!30} w/ ATII (Ours) & 4.48M & $\textbf{0.040}$ & $\textbf{0.511}$ & $\textbf{0.795}$ \\ \toprule
\end{tabular}
\caption{\textbf{Evaluation of Different Scans in Mamba and Analysis on Adaptive Textual Information Injectior (ATII).} SDS denotes the original single-directional scan in Mamba. BDS denotes the bidirectional scan in \cite{vimamba}. PBDS denotes our pseudo-bidirectional scan. ``w/o gating'' denotes that we directly feed the concatenation of segment token and text token into the fusion layer.}
\label{tab:scan}
\end{table}

\section{Conclusion}
In this work, we introduce Light-T2M, a lightweight and fast model for text-to-motion generation that underscores the importance and effectiveness of local information modeling. Furthermore, Light-T2M incorporates the Mamba and pseudo-bidirectional scan for global information modeling, both of which significantly reduce parameters. Light-T2M features an adaptive textual information injector, an efficient way to control the generated motion through text. Through these designs, Light-T2M achieves superior performance while using only a small number of parameters.

\section{Acknowledgments}
This work was supported partially by NSFC(92470202, U21A20471) and Guangdong NSF Project (No. 2023B1515040025). 

\bibliography{aaai25}
\clearpage

\appendix

\section{Appendix}

In this appendix, we provide additional details about our Light-T2M. We also provide a supplementary video for dynamic visualizations. In the following, we first introduce more details of our model design. Then, we illustrate more implementation details. We also provide more experiments to demonstrate the effectiveness of our method.  Finally, we include a discussion on a concurrent work, Motion Mamba.

\section{More Details about Light-T2M}

\subsection{Training and Inference Process}
As mentioned in our main paper, our training target is predicting sample $M^0$ instead of noise $\epsilon$. As noted in DDPM \cite{ddpm}, they are mathematically equivalent. Given $M^t$ and $M^0$, we can calculate noise $\epsilon$ via
\begin{equation}
    \epsilon = \frac{M^t - \sqrt{\bar{\alpha_t}}M^0}{\sqrt{1-\bar{\alpha_t}}}. \label{eq:noise}
\end{equation}
During sampling, given $M^t$, the conditional sample $\hat{M}_*^0$ and the unconditional sample $\bar{M}_*^0$ can be predicted by setting $c$ as the given text $W$ and $\varnothing$. Then, to use the \textit{classifier-free guidance} method \cite{cfg} to achieve conditional generation, we first calculate the conditional noise $\epsilon_c$ and unconditional noise $\epsilon_u$ via Equation \ref{eq:noise}. After that, we implement the \textit{classifier-free guidance} method via Equation 2 (in the main paper) to obtain predicted $\hat{\epsilon}$. With the predicted $\hat{\epsilon}$ and noise sample $M^t$, we can calculate $M^{t-1}$.

\section{More Implementation Details}
During training, we test on the validation set and save a checkpoint after some epochs, and we select the checkpoint with the best FID on the validation set as the final model. In addition to the content introduced in the main paper, we provide further details in Table \ref{tab:hyperparameters}. For the Mamba block, the SSM state expansion factor, local convolution width, and block expansion factor are 16, 4, and 2, respectively. The kernel size and the stride in Depth-wise Conv1D are 3 and 1.

\begin{algorithm}[]
\caption{Training}
\label{alg:algorithm}
\begin{algorithmic}[1] 
\REPEAT
\STATE sample motion $M^0$ and text $W$ from the dataset
\STATE t $\sim$ Uniform({1, ..., T})
\STATE $\epsilon \sim \mathcal{N}(\mathbf{0}, \mathbf{I})$
\STATE Set $c$ to $\varnothing$ or $W$ with probability $\tau$
\STATE $M^t = \sqrt{\bar{\alpha}_t}M^0 + \sqrt{1-\bar{\alpha_t}}\epsilon$
\STATE Take gradient descent step on \\
$\quad\quad \nabla _\theta || M^0 - \phi_\theta(M^t, t, c) ||^2 $ 
\UNTIL{converged} 
\end{algorithmic}
\end{algorithm}

\begin{algorithm}[]
\caption{Sampling}
\label{alg:algorithm}
\begin{algorithmic}[1] 
\STATE $M^T \sim \mathcal{N}(\mathbf{0}, \mathbf{I})$
\FOR{$t=T,...,1$}
\STATE $\hat{M}_*^0 = \phi_\theta(M^t, t, W)$
\STATE $\bar{M}_*^0 = \phi_\theta(M^t, t, \varnothing)$
\STATE Calculate $\epsilon_c$ and $\epsilon_u$ via Eq. \ref{eq:noise}
\STATE Calculate $\hat{\epsilon}$ via Eq. 2
\STATE Calculate $M^{t-1}$ using $M^{t}$ and $\hat{\epsilon}$
\ENDFOR 
\end{algorithmic}
\end{algorithm}

\begin{table}[]
    \centering
    \begin{tabular}{c|c}
    \toprule
        Hyperparameter &  \\ \hline
         Optimizer & AdamW \\
         Learning rate & 2e-4  \\
         Learning rate scheduler & cosine \\
         $\beta_1$ & 0.9 \\
         $\beta_2$ & 0.999\\
         Batch size & 256 \\ \hline
         Guidance scale $s$ & 4 \\
         Text dropout ratio $\tau$ & 0.2 \\
         \hline
         Model hidden dim  $D$ & 256 \\
         Number of basic blocks $N$ & 4\\
         Downsampling scale $S$ & 8 \\
         Number of Groups in GroupNorm & 16\\
         Stride and padding in 1D CNN & 3, 1 \\
         State expansion factor in Mamba & 16 \\ 
         Local convolution width in Mamba & 4 \\ 
         Block expansion factor in Mamba & 2 \\ \toprule
    \end{tabular}
    \caption{\textbf{Hyperparameters for the HumanML3D and KIT-ML datasets.}}
    \label{tab:hyperparameters}
\end{table}


\section{Additional Experiments}

\subsection{Details of Inference Time Evaluation}
As introduced in the main paper, Average Inference time (AIT) is calculated from the average across 100 samples using the same RTX 3090Ti GPU.  To obtain more accurate results, we replicated the experiment 20 times to calculate the average AIT. For all methods, we do not use the mixed precision during generation. We first randomly select 100 samples from the test set. Then, during evaluation, these selected samples are used for all methods (the test samples used are the same across all methods).

\begin{table}[]
\centering
\setlength\tabcolsep{1.5mm}
\begin{tabular}{ccccc}
\bottomrule
Block Design & \#Params & FID$\downarrow$  & R-Top1$\uparrow$ & R-Top3$\uparrow$ \\ \toprule
TTT & 9.81M & 0.175 & 0.472 & 0.766\\
T & {3.49M} & 0.214 & 0.437 & 0.727\\
LTL & 4.03M & 0.160 & 0.469 & 0.763 \\
LML & $\textbf{2.63M}$ & 0.165 & 0.456 & 0.736\\
LM$^*$L & 4.38M & ${0.114}$ & ${0.501}$ & ${0.783}$\\
\rowcolor{gray!30} LGL (Ours) & 4.48M & $\textbf{0.040}$ & $\textbf{0.511}$ & $\textbf{0.795}$ \\ \toprule
G & \textbf{3.94M} & 0.158 & 0.490 & 0.778 \\
LG & 4.21M & 0.076 & 0.507 & 0.789\\
LLG & 4.48M &  0.042 & 0.510 & 0.791\\
GL & 4.21M  & 0.052 & 0.505 & 0.790\\
GLL & 4.48M & 0.043 & 0.510 & 0.792\\
\rowcolor{gray!30} LGL (Ours)  & 4.48M & $\textbf{0.040}$ & $\textbf{0.511}$ & $\textbf{0.795}$ \\ \toprule
\end{tabular}
\caption{\textbf{Analysis of Model Design.} We evaluate the performance when replacing basic blocks in Light-T2M with other basic blocks. ``T'' denotes one Transformer encoder layer. ``L'' denotes our Local Information Modeling Module. ``G'' denotes our Global Information Modeling and Injection Textual Information Module. ``M'' and ``M$^*$'' denote an original Mamba block and a BiMamba block in \cite{vimamba}.}
\label{tab:model}
\end{table}

\subsection{More Experiments about Model Design}
We conducted additional experiments to assess the performance of various configurations combining the Local Information Modeling Module with the Global Information Modeling and Textual Information Injection Module. The findings, detailed in Table \ref{tab:model}, indicate that incorporating the Local Information Modeling Module enhances the model's performance, underscoring its significance. Comparisons among configurations ``LG'', ``LLG'', ``GL'', ``GLL'', and ``LGL'' reveal that the ``LGL'' design yields the best results.

We note that although the design of ``TTT'' seems similar to that of MDM \cite{mdm}, our ``TTT'' and MDM have significant differences in many details, which enable us to train a purely transformer-based model to achieve high performance (an FID of 0.175). Here, we list some of the important differences: (1) we treat the diffusion timestep as a separate token, rather than integrating it with the text embedding token; (2) the details of the Transformer differ, including the number of channels, layers, heads, and the dimension expansion factor in the FFN layer; (3) the number of training epochs, batch size, learning rate, and learning rate decay strategy we use are all different.

\subsection{Impact of Guidance Scale}
As shown in Table \ref{tab:gd}, we evaluate the impact of the guidance scale when using the \textit{classifier-free guidance} method. As the guidance scale increases from 1 to 4, our model's performance gradually improves. However, when the guidance scale reaches 5 or higher, the performance begins to decline rapidly. Our model achieves excellent performance when the guidance scale is set to either 3 or 4.

\subsection{Impact of Downsampling Scale}
As introduced in the main paper, the downsampling has two benefits. As shown in Table \ref{tab:gd}, we evaluate the impact of the downsampling scale. When the downsampling scale is set to 1, our model achieves a respectable FID, but the R-Top 1 is relatively low. However, when the downsampling scale is increased to 8, our model performs optimally. Subsequently, as the downsampling scale continues to increase to 16, our model's performance slightly decreases, though the decline is not significant. These results demonstrate the benefits of downsampling discussed in our paper.

\begin{table}[]
\centering
\begin{tabular}{cccc}
\bottomrule
Downsampling Scale & FID$\downarrow$  & R-Top1$\uparrow$ & R-Top3$\uparrow$\\ \toprule
1 & 0.060 & 0.501 & 0.774\\
4 & 0.045 & 0.505 & 0.779\\
\rowcolor{gray!30} 8 & 0.040 & 0.511 & 0.795\\
16 & 0.043 & 0.509 & 0.791\\ \toprule
\end{tabular}
\caption{\textbf{Evaluation of Different Downsampling Scales.} }
\label{tab:ds}
\end{table}

\begin{table}[t]
\centering
\begin{tabular}{cccc}
\bottomrule
Sampling Step & FID$\downarrow$  & R-Top1$\uparrow$ & R-Top3$\uparrow$\\ \toprule
 \multicolumn{4}{c}{\textit{Using UniPC.}} \\ \hline
5 & 0.226 & 0.425 & 0.703\\
7 & 0.145 & 0.475 & 0.756\\
9 & 0.042 & 0.508 & 0.792 \\
\rowcolor{gray!30} 10 & 0.040 & 0.511 & 0.795\\
15 & \textbf{0.039} & 0.511 & 0.794\\
20 & \textbf{0.039} & \textbf{0.512} & \textbf{0.796}\\ \toprule
 \multicolumn{4}{c}{\textit{Using DDIM.}} \\ \hline
5 & 0.083 & 0.506 & 0.790 \\
10 & 0.061 & 0.509 & \textbf{0.794} \\
20 & 0.053 & \textbf{0.510} & \textbf{0.794} \\
30 & 0.046 & 0.509 & 0.792 \\
40 & 0.043 & 0.508 & 0.791 \\
50 & \textbf{0.041} & \textbf{0.510} & 0.793 \\ \toprule
\end{tabular}
\caption{\textbf{Evaluation of Sampling Step and Sample Scheduler.} }
\label{tab:ds}
\end{table}

\begin{table*}[]
\centering
\setlength\tabcolsep{1mm}
\begin{tabular}{ccccccccc}
\bottomrule
 \multirow{2}{*}{Methods} & \multirow{2}{*}{\#Params} & \multirow{2}{*}{AIT$\downarrow$} & \multirow{2}{*}{FID$\downarrow$}  & \multicolumn{3}{c}{R-Precision$\uparrow$} & \multirow{2}{*}{MM. Dist.$\downarrow$} & \multirow{2}{*}{MM. $\uparrow$} \\ \cline{5-7}
   & & & & Top1 & Top2 & Top3 & & \\ \toprule
\rowcolor{gray!30} \multicolumn{9}{l}{\textit{On the HumanML3D dataset.}}\\ \hline 
Motion Mamba (\citeauthor{motionmamba}) & -& - & $0.281^{\pm.009}$ & $0.502^{\pm.003}$ & $0.693^{\pm.002}$ & $0.792^{\pm.002}$ & $3.060^{\pm.058}$ & $\textbf{2.294}^{\pm.058}$ \\ \hline
\textbf{Our Light-T2M} & \textbf{4.48M} & \textbf{0.151s} & $\textbf{0.040}^{\pm{.002}}$ & $\textbf{0.511}^{\pm.003}$ & $\textbf{0.699}^{\pm.002}$ & $\textbf{0.795}^{\pm.002}$ & $\textbf{3.002}^{\pm.008}$ & $1.670^{\pm.061}$ \\ \toprule

\rowcolor{gray!30} \multicolumn{9}{l}{\textit{On the KIT-ML dataset.}}\\ \hline 
Motion Mamba (\citeauthor{motionmamba}) & - & - & $0.307^{\pm.041}$ & $0.419^{\pm.006}$ & $0.645^{\pm.005}$ & $0.765^{\pm.006}$ & $3.021^{\pm.025}$ & $\textbf{1.678}^{\pm.064}$ \\ \hline
\textbf{Our Light-T2M} & - & - & $\textbf{0.161}^{\pm.009}$ & $\textbf{0.444}^{\pm.006}$ & $\textbf{0.670}^{\pm.007}$ & $\textbf{0.794}^{\pm.005}$ & $\textbf{2.746}^{\pm.016}$ & $1.005^{\pm.036}$\\ \toprule
\end{tabular}
\caption{\textbf{Quantitative comparison with Motion Mamba on the HumanML3D and KIT-ML test set.} } 
\label{tab:hml}
\end{table*}

\begin{table}[]
\centering
\begin{tabular}{cccc}
\bottomrule
Guidance Scale & FID$\downarrow$  & R-Top1$\uparrow$ & R-Top3$\uparrow$\\ \toprule
1 & 0.322 & 0.423 & 0.708\\
2 & 0.079 & 0.492 & 0.777\\
3 & 0.041 & 0.509 & 0.792\\
\rowcolor{gray!30} 4 & 0.040 & 0.511 & 0.795\\
5 & 0.052 & 0.501 & 0.789\\
6 & 0.105 & 0.490 & 0.769\\ \toprule
\end{tabular}
\caption{\textbf{Evaluation of Different Guidance Scales.} }
\label{tab:gd}
\end{table}

\subsection{Impact of Sampling Step}
As shown in Table \ref{tab:gd}, we further evaluate the impact of the sample step when using UniPC \cite{unipc} or DDIM \cite{ddim}. With UniPC, setting the sampling steps to 5 leads to a sharp decline in performance, with an FID of 0.226. However, a slight increase in sampling steps significantly enhances performance. Beyond 10 steps, additional improvements are minimal and result in longer inference times. Interestingly, at 5 steps, the model performs better with DDIM than with UniPC. Yet, as the number of steps increases, the performance using UniPC substantially surpasses that of using DDIM.

\section{Discussion about Motion Mamba}
We acknowledge that Motion Mamba, a Mamba-based network for text-to-motion generation, is \textbf{a concurrent work}. Significantly, our model surpasses Motion Mamba by a considerable margin, as detailed in Table \ref{tab:hml}. Additionally, since the source code for Motion Mamba is not publicly available, we could not assess its average inference time.

\end{document}